\pdfoutput=1
\documentclass[11pt]{article}
\usepackage[preprint]{acl}

\usepackage[utf8]{inputenc}
\usepackage{times}
\usepackage{latexsym}

\usepackage[T1]{fontenc}

\usepackage{microtype}
\usepackage{inconsolata}
\usepackage{graphicx}

\usepackage{amssymb}
\usepackage{amsmath}
\usepackage{booktabs}
\usepackage{array}
\usepackage{multirow}
\usepackage{tabularx}
\usepackage{hyperref}
\usepackage{tcolorbox}
\usepackage{CJKutf8}
\usepackage{subfigure}
\usepackage{hyperref}
\usepackage{colortbl}

\title{Dynamic Chunking and Selection for Reading Comprehension of Ultra-Long Context in Large Language Models}

\author{
    \textbf{Boheng Sheng\textsuperscript{1}},
    \textbf{Jiacheng Yao\textsuperscript{1}},
    \textbf{Meicong Zhang\textsuperscript{1}},
    \textbf{Guoxiu He\textsuperscript{1}\thanks{Corresponding author}},
    \\
    \textsuperscript{1}East China Normal University,
    \\
    \texttt{\{bhsheng,jcyao,mczhang\}@stu.ecnu.edu.cn} \\
    \texttt{gxhe@fem.ecnu.edu.cn}
    \\
}

\begin{document}
\begin{CJK}{UTF8}{gbsn}
\maketitle
\begin{abstract}
Large language models (LLMs) often struggle to accurately read and comprehend extremely long texts. Current methods for improvement typically rely on splitting long contexts into fixed-length chunks. However, fixed truncation risks separating semantically relevant content, leading to ambiguity and compromising accurate understanding. To overcome this limitation, we propose a straightforward approach for dynamically separating and selecting chunks of long context, facilitating a more streamlined input for LLMs. In particular, we compute semantic similarities between adjacent sentences, using lower similarities to adaptively divide long contexts into variable-length chunks. We further train a question-aware classifier to select sensitive chunks that are critical for answering specific questions. Experimental results on both single-hop and multi-hop question-answering benchmarks show that the proposed approach consistently outperforms strong baselines. Notably, it maintains robustness across a wide range of input lengths, handling sequences of up to 256k tokens. Our datasets and code are available at the following link: \href{https://github.com/ECNU-Text-Computing/DCS}{https://github.com/ECNU-Text-Computing/DCS}. 

\end{abstract}

\section{Introduction}
\label{sec:intro}

\begin{figure}[t]
  \centering
  \includegraphics[width=0.99\linewidth]{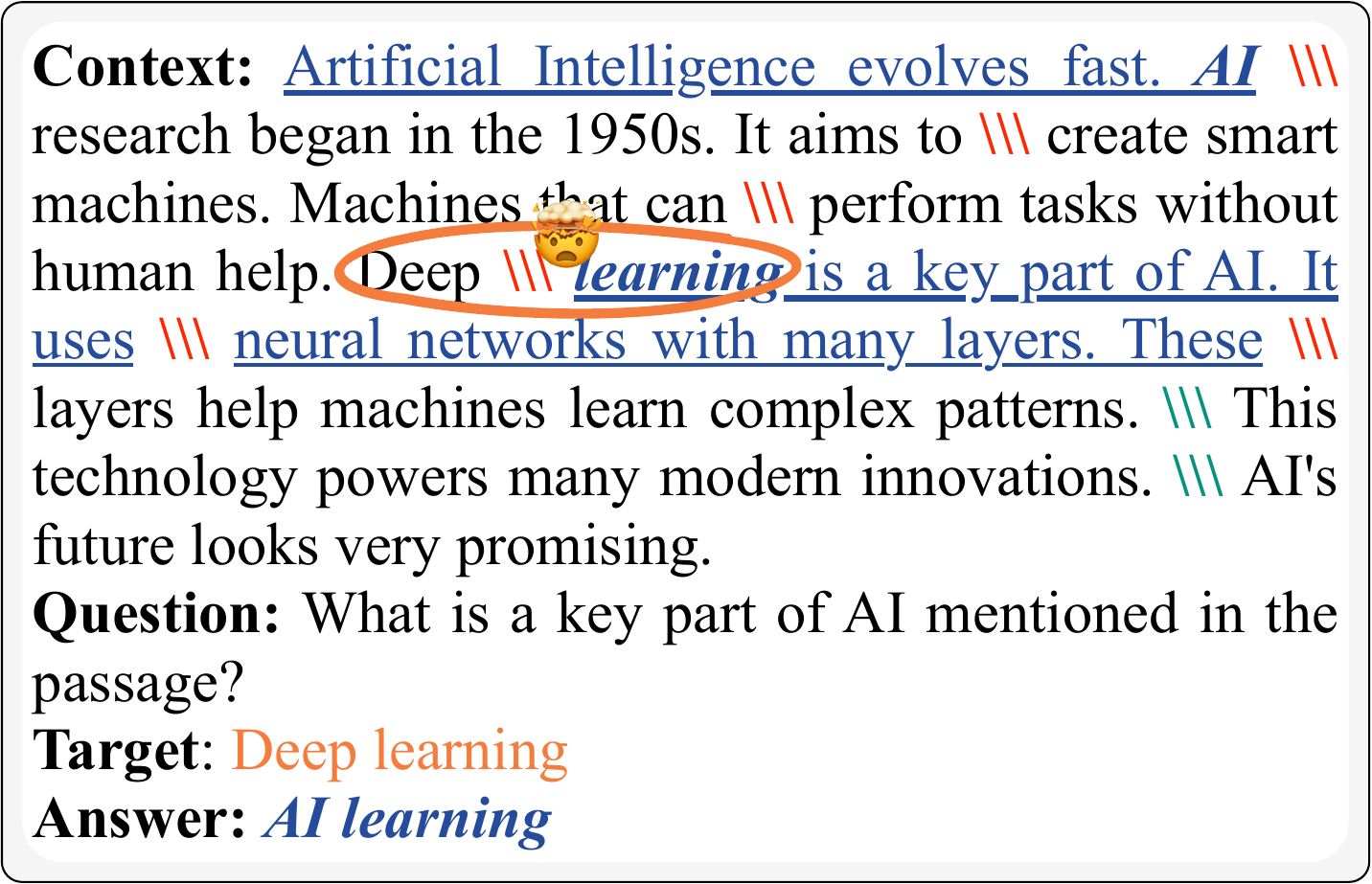}
  \caption{A failure case of a fixed-length chunking method in the QA task. The context is segmented into fixed-length chunks, with \textbackslash\textbackslash\textbackslash~ indicating the split points. The \textcolor[HTML]{2C4D96}{\underline{blue underlines}} highlight the chunks selected by the method as relevant. However, the key phrase \textcolor[HTML]{E6824C}{Deep learning} is split across two separate chunks, preventing the LLM from capturing its full meaning. As a result, the LLM produces an incorrect answer.}
  \label{intro_fig}
\end{figure}

Recent advances in large language models (LLMs) \cite{openai2024gpt4technicalreport,touvron2023llamaopenefficientfoundation,touvron2023llama2openfoundation,bai2023qwentechnicalreport} have revolutionized the landscape of natural language processing (NLP), demonstrating remarkable capabilities in various tasks such as machine translation \cite{lu2024llamaxscalinglinguistichorizons,xu2024contrastivepreferenceoptimizationpushing}, text summarization \cite{tam-etal-2023-evaluating,zhang-etal-2024-benchmarking}, and reading comprehension \cite{samuel-etal-2024-llms}. While LLMs are designed to process long texts, they still encounter challenges in achieving accurate understanding in real-world applications \cite{10.1162/tacl_a_00638}. This issue is particularly evident when LLMs answer specific questions based on very lengthy texts.

On the one hand, there are inherent flaws in the pre-trained Transformer Decoder architecture \cite{wang2024limitssurveytechniquesextend}. Notably, the scope of positional encoding limits the input context window to a fixed length; the quadratic attention computational complexity constrains input length based on available computational resources. On the other hand, empirical studies show that LLMs tend to disproportionately allocate attention to the beginning and end of input \cite{liu2023lostmiddlelanguagemodels}. Therefore, when question-sensitive information is located in the middle, LLMs often fail to incorporate these critical details into their answer generation. These limitations lead to poor performance, driving the development of methods that efficiently enhance the long-context understanding capabilities of LLMs.

Intuitive improvements hinge on breaking lengthy text into manageable pieces and applying targeted operations to them to enhance the adaptability of LLMs to long texts \cite{xiao2024infllm,song2024hierarchical,an2024trainingfreelongcontextscalinglarge}.
However, current methods often only divide the input into fixed-length chunks, which can severely compromise semantic coherence. As shown in Figure \ref{intro_fig}, when the input context is segmented by fixed lengths, breakpoints frequently occur in the middle of sentences, resulting in only a small portion of sentences being fully preserved within a single chunk. First of all, this fragmentation undermines the logical structure of the original text, making it difficult to grasp the semantic connections between chunks during the selection process. This can hinder overall comprehension of the context. Moreover, if a sentence contains crucial information or answers, fragmentation risks distorting its meaning, leading to the exclusion of related sentences and resulting in inaccurate responses. To address this issue, it is essential to dynamically determine chunking boundaries based on semantic structure and flexibly select the most relevant chunks.

In this paper, we propose a straightforward approach for LLMs, termed Dynamic Chunking and Selection (DCS). This approach aims to effectively tackle the challenge of reading comprehension within extensive contexts. In particular, we utilize Sentence-BERT \cite{reimers-2019-sentence-bert} to encode lengthy context at the sentence level. Then, by assessing the semantic similarity among adjacent sentences, we dynamically segment the context into variable-length chunks. This ensures that each chunk retains its inherent coherence and semantic integrity. Next, we train a question-aware classifier to select chunks based on the provided question. This classifier rigorously evaluates the relevance of each chunk to the question, selecting only those that contain essential information. This process allows LLMs to preserve maximum relevant content while adhering to length constraints. Finally, the selected chunks are concatenated in their original order and fed into the LLM. The conciseness and comprehensiveness of the input enable the LLM to generate accurate responses while maintaining the integrity of the original narrative structure. As a result, this approach could enhance the LLM's ability to process and understand extensive contexts.

To evaluate the performance of our approach, we conduct comprehensive experiments based on three base LLMs: Llama-3-8B-Instruct \cite{grattafiori2024llama3herdmodels}, Mistral-7B-Instruct \cite{jiang2023mistral7b}, and Vicuna-7B \cite{zheng2023judgingllmasajudgemtbenchchatbot}. Our evaluation encompasses 12 diverse long-context reading comprehension datasets, covering both single-hop and multi-hop question-answering (QA) tasks. To further scrutinize our approach's capabilities, we also test it on significantly longer datasets (up to 256k tokens). The results demonstrate that our approach consistently outperforms recent state-of-the-art (SOTA) methods across most datasets. Moreover, experiments on ultra-long texts underscore our approach's robustness and potential for effectively handling extensive contexts.

In summary, our main contribution is the introduction of Dynamic Chunking and Selection (DCS). This approach is both straightforward and highly effective, addressing the challenges of long-context reading comprehension without requiring complex architectures. DCS involves Sentence-BERT for sentence embeddings, dynamically segments texts based on semantic similarity, and utilizes a question-aware classifier to select relevant chunks. This minimalist design ensures ease of implementation and minimal training overhead while achieving significant performance improvements. Our approach offers a reliable and efficient solution for LLMs dealing with extensive contexts.

\section{Related Work}
\label{sec:relate}

Since the emergence of LLMs, extensive research has focused on enabling them to process longer contexts.

\noindent\textbf{Context Length Extrapolation.} 
\citet{chen2023extending} introduced Position Interpolation (PI), a methodology that expands the context window dimensions of RoPE-based LLMs \cite{su2024roformer} while maintaining relative positional relationships. 
Subsequent developments such as YaRN \cite{peng2023yarn} demonstrate superior performance compared to existing RoPE interpolation approaches. This optimized technique serves as a direct substitute for PI implementations while substantially expanding their applicability, maintaining backward compatibility with existing architectures. 
However, these methods only address the issue of long input. They do not fully address the challenge of LLMs in capturing long-context dependencies.

\noindent\textbf{Sparse Attention.} 
StreamingLLM \cite{xiao2023efficient} employs a dual-component architecture combining sliding-window attention with attention-sink mechanisms, enabling stable processing of arbitrarily long text sequences without model retraining. 
LM-Infinite \cite{han2024lm} implements two elements: a $\Lambda$-shaped attention mask for gradient stabilization and a distance ceiling parameter, while strategically reintroducing intermediate top-k tokens to optimize downstream task performance. 
Longformer \cite{beltagy2020longformerlongdocumenttransformer} employs a linearly scaling attention mechanism combining local and global attentions. This enables efficient processing of lengthy documents. 
Although sparse attention mechanisms can enhance the ability of LLMs to comprehend long contexts. Their reliance on predefined methods to reduce the computational cost of attention inevitably limits the potential for significant performance improvements.

\noindent\textbf{Tokens Eviction.} 
Heavy Hitter Oracle (H2O) \cite{zhang2023h2oheavyhitteroracleefficient} introduces a novel KV cache eviction policy. It identifies and retains "Heavy Hitter" tokens that significantly contribute to attention scores. By dynamically balancing recent and critical tokens, H2O can comprehend long inputs. Token Omission Via Attention (TOVA) \cite{oren2024transformersmultistaternns} is another training-free compression policy for reducing the key-value cache size. By conceptualizing decoder-only transformers as unbounded multistate RNNs, TOVA uses in some cases only 1/8 of the original cache size to handle longer sequences. Chunked Instruction-aware State Eviction (CItruS) \cite{bai2024citruschunkedinstructionawarestate} integrates attention preferences relevant to downstream tasks into the eviction process. It improves performance on long sequence comprehension and retrieval tasks while maintaining language modeling perplexity. Token eviction methods effectively balance model performance and resource usage. However, they fail to fully preserve the original semantic structure of the text, thereby constraining potential performance improvements. In contrast, our chunk-level approach effectively addresses this limitation.

\noindent\textbf{Chunk-level Processing.}
InfLLM \cite{xiao2024infllm} addresses memory constraints through distributed context storage, utilizing specialized memory units with content-aware indexing for efficient retrieval during attention computations.
Hierarchical Memory Transformer (HMT) \cite{he2024hmthierarchicalmemorytransformer} establishes a biologically-inspired architecture that emulates human memory organization through multi-granular memory consolidation. The framework employs pyramidal memory cells with differential retention policies. It also combines segment-level recurrence with content-based memory reactivation to maintain coherent long-range dependencies.
However, the methods mentioned above segment the text into fixed lengths, potentially undermining the semantic integrity of the original text. In contrast, our approach employs a dynamic segmentation to preserve the semantic coherence of the input text.

\noindent\textbf{Tuning based Methods.}
Building upon Low-Rank Adaptation (LoRA) \cite{DBLP:journals/corr/abs-2106-09685}, \citet{chen2024longloraefficientfinetuninglongcontext} devised LongLoRA, which combines modified sparse attention patterns with optimized low-rank decomposition strategies to efficiently extend LLMs' context processing capacity while preserving computational frugality.
Unlimiformer \cite{bertsch2023unlimiformerlongrangetransformersunlimited} involves a memory-efficient adaptation strategy. It enables the processing of arbitrarily long sequences through context-aware clustering with cross-attention. 
MEGALODON \cite{ma2024megalodonefficientllmpretraining} presents an efficient neural architecture framework for unbounded sequence modeling. This architecture incorporates three core components: complex exponential moving average operators for temporal dependency modeling, learnable timestep normalization layers, and enhanced attention mechanisms with adaptive span control.
Although these methods can achieve satisfactory results, they require extensive training that consumes significant computational resources, both in terms of space and time. In contrast, our approach achieves substantial improvements in model performance with minimal training overhead.

\begin{figure*}[t]
  \centering
  \includegraphics[width=0.85\linewidth]{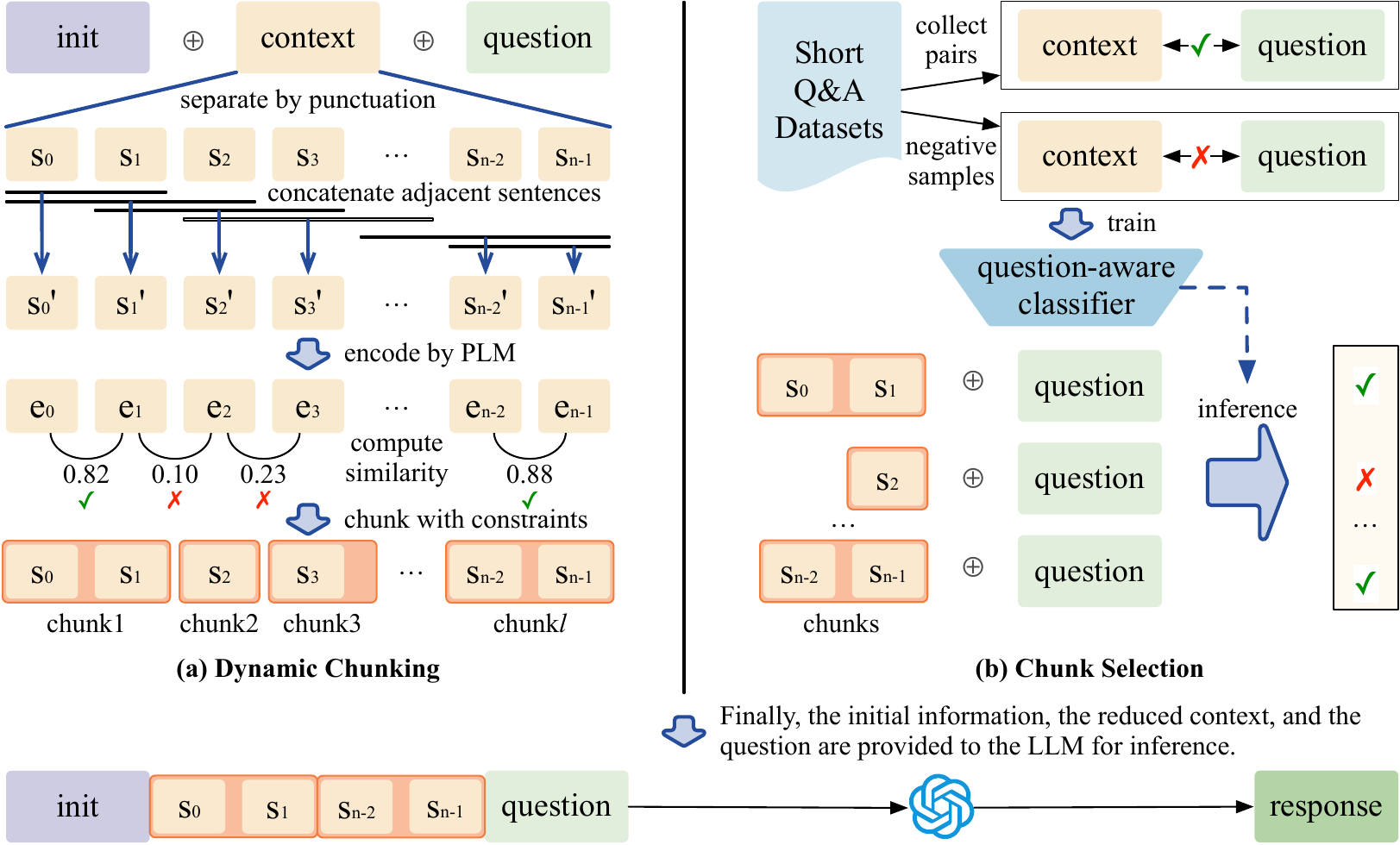}
  \caption {The overall structure of the proposed DCS. It includes two small modules to compress the input to help the LLM understand long context better and derive the correct answer.}
  \label{model}
\end{figure*}
 
\section{Methodology}
\label{sec:method}

This section introduces Dynamic Chunking and Selection (DCS) for LLMs towards reading comprehension. DCS dynamically segments long-context inputs into discrete chunks. Then it meticulously filters out irrelevant text fragments. After that, it concatenates the remaining text to fit within the predefined context window constraints of LLMs. This methodology significantly enhances the ability of LLMs to process contextual information effectively. The overall structure of DCS is shown in Figure \ref{model}. We also place a notation table as shown in Table \ref{Notation Table} to facilitate readers' reference.

\subsection{Dynamic Chunking}
Our approach initiates with semantic segmentation \cite{FullStackRetrieval.com}
applied to input context $C$, structured through three components: [initial information, context, question]. The context component undergoes punctuation-driven decomposition, generating sentence sequence $[s_0, s_1, \cdots, s_{n-1}]$ where $n$ denotes total sentence count. 

To preserve the semantic integrity of individual sentences when they are separated from the broader context, it is necessary to concatenate adjacent sentences before encoding them. Specifically, given the predefined chunk length parameter $l$, contextual expansion is performed via neighborhood merging:
\begin{equation}
s'_i = 
\begin{cases}
s_0 \oplus s_1 & i = 0, \\
s_{i-1} \oplus s_i \oplus s_{i+1} & 1 \leq i \leq n-2, \\
s_{n-2} \oplus s_{n-1} & i = n-1,
\end{cases}
\end{equation}
yielding enhanced context segments $[s'_0, s'_1, \cdots, s'_{n-1}]$.

The merged segments undergo encoding via pre-trained sentence-BERT to obtain contextual embeddings $[e_0, e_1, \cdots, e_{n-1}] \in R^d$. Adjacent embedding pairs then undergo similarity measurement through cosine similarity computation:  
\begin{equation}
\text{sim}(i, i+1) = \frac{e_i^\top e_{i+1}}{\|e_i\| \|e_{i+1}\|},
\end{equation}  
where similarity scores monotonically increase with semantic congruence. For boundary detection between context chunks, the semantic dissimilarity metric is derived through cosine distance transformation:  
\begin{equation}
\text{dis}(i) = 1 - \text{sim}(i, i+1).
\end{equation}  

The semantic cosine distance sequence $[dis_0, dis_1, \cdots, dis_{n-2}]$ undergoes ascending-order sorting to produce ordered indices $[k_0, k_1, \cdots, k_{n-2}]$ where $dis_{k_0} \leq dis_{k_1} \leq \cdots \leq dis_{k_{n-2}}$. A percentile-based segmentation threshold $\alpha \in [0,1]$ determines boundary selection through quantile computation:  
\begin{equation}
\mathcal{K} = \left[ k_{\lceil (1-\alpha) n \rceil}, \cdots, k_{n-2} \right],
\end{equation}
which preserves the top $(1-\alpha)$ proportion of maximal dissimilarity indices as segmentation boundaries. The original document $C$ is partitioned at positions $\mathcal{K}$ through binary splitting, generating final document segmentation:  
\begin{equation}
\mathcal{C} = \left[ c^{(0)}_0, c^{(0)}_1, \cdots, c^{(0)}_{m_0} \right], \quad m_0 = |\mathcal{K}|.
\end{equation}
  
The segmentation refinement phase ensures compliance with pre-specified chunk length constraint $l$ through iterative optimization. The initial segmentation $\mathcal{C}^{(0)}$ undergoes recursive reprocessing until iteration $j$ where $\max_{k} |c^{(j)}_k| > l$ triggers termination. The preceding iteration's output $\mathcal{C}^{(j-1)} = [c^{(j-1)}_0, \cdots, c^{(j-1)}_{m_{j-1}}]$ is selected as baseline segmentation. Given the current significant variability in chunk sizes, further merging of the blocks is performed to make each chunk as close as possible to the predefined chunk size $l$. Specifically, for each starting chunk $ c_i $, find the smallest integer $ u $ such that: 
\begin{align}
  &\sum_{j=i}^{i+u} |c_j| \leq l \Rightarrow c_i \oplus \cdots \oplus c_{i+u}, \nonumber \\
  & c_i, \cdots, c_{i+u} \in \mathcal{C}^{(j-1)}.
\end{align}
 
After merging, we update the index $ i $ to $ i + u + 1 $ and continue processing the next unmerged chunk and yield final chunks $\mathcal{C} = [c_0, \cdots, c_m]$ with $|c_k| \leq l, \forall k \leq m$. The processed document structure maintains the original framing components:  
\begin{equation}
C_{\text{processed}} = [\text{initial}, \mathcal{C}, \text{question}].
\end{equation}

\subsection{Chunk Selection}
A question-aware classification model is subsequently trained to optimize chunk selection through question-relevance assessment. 

\noindent\textbf{Training Data Collection.} The training data is curated from question-answering corpora with controlled complexity and scale. Authentic context-question pairs $[C,Q]$ are extracted as positive training samples through exhaustive enumeration. Complementary negative samples are generated via negative sampling strategy $\mathcal{S}: \mathcal{D} \rightarrow \mathcal{D}^-$, where $\mathcal{D}$ denotes original dataset and $\mathcal{D}^-$ represents semantically uncorrelated pairs. For each processed pair $[C,Q]$, context and question tokens are concatenated into a unified sequence:  
\begin{equation}
X = [C_0, \cdots, C_{p-1}; Q_0, \cdots, Q_{q-1}] \in \mathbb{N}^{(p+q) \times d},
\end{equation}
where $p = |C|$ and $q = |Q|$ denote sequence length. This composite sequence is encoded through the LLM's transformer layers, producing final-layer representations:  
\begin{equation}
H = [h_0^{(d)}, h_1^{(d)}, \cdots, h_{p+q-1}^{(d)}] \in \mathbb{R}^{(p+q) \times d},
\end{equation}
and multi-head attention scores:  
\begin{equation}
\mathcal{A} \in \mathbb{R}^{n_h \times n_l \times n_l} \quad (n_l = p+q, \ d \in \mathbb{N}^+),
\end{equation} 
where $n_h$ indicates the number of parallel attention heads.

Utilizing complete sequence encodings $H \in \mathbb{R}^{n_l \times d}$ for classifier training induces prohibitive computational complexity $\mathcal{O}(n_l^2)$. To mitigate this, we implement feature distillation through strategic state selection from the final transformer layer. The extraction protocol first captures boundary tokens:  
\begin{equation}
H_b = [h^{(d)}_0, h^{(d)}_{p-1}, h^{(d)}_p, h^{(d)}_{p+q-1}].
\end{equation} 
And the attention scores are averaged along the head dimension:
\begin{align}
\mathcal{A}_h &= \tfrac{1}{n_h} \sum_{i=0}^{n_h-1} \mathcal{A}_i \in \mathbb{R}^{n_l \times n_l}
\end{align}  

Then the attention matrix $\mathcal{A}_h \in \mathbb{R}^{n_l \times n_l}$ is decomposed into four submatrices through block partitioning:
\begin{equation}
\mathcal{A}_h = 
\left[
\begin{array}{c|c}
\mathcal{A}_{CC} \in \mathbb{R}^{p \times p} & \mathcal{A}_{CQ} \in \mathbb{R}^{p \times q} \\
\hline
\mathcal{A}_{QC} \in \mathbb{R}^{q \times p} & \mathcal{A}_{QQ} \in \mathbb{R}^{q \times q}
\end{array}
\right],
\end{equation}
where $\mathcal{A}_{QC}$ captures cross-attention between question tokens and context tokens (Q→C), while $\mathcal{A}_{QQ}$ represents intra-attention within question tokens (Q$\to$Q). Column-wise mean pooling is applied to both submatrices: % →
\begin{align}
\mathbf{a}_C &= \frac{1}{q} \sum_{j=1}^q \mathcal{A}_{QC}(j,:) \in \mathbb{R}^p, \\
\mathbf{a}_Q &= \frac{1}{q} \sum_{j=1}^q \mathcal{A}_{QQ}(j,:) \in \mathbb{R}^q.
\end{align}

These attention weights are then used to compute context-specific and question-specific representations:
\begin{align}
h^{(d)}_C &= \mathbf{a}_C \cdot [h^{(d)}_0, \cdots, h^{(d)}_{p-1}]^\top \in \mathbb{R}^d, \\
h^{(d)}_Q &= \mathbf{a}_Q \cdot [h^{(d)}_p, \cdots, h^{(d)}_{p+q-1}]^\top \in \mathbb{R}^d.
\end{align}

The final feature matrix concatenates boundary tokens with attention-pooled vectors:
\begin{equation}
H = [h^{(d)}_0; h^{(d)}_C; h^{(d)}_{p-1}; h^{(d)}_{p}; h^{(d)}_Q; h^{(d)}_{p+q-1}] \in \mathbb{R}^{6 \times d},
\end{equation}
which serves as the classifier input tensor.

\noindent\textbf{Classifier Training.} The classifier employs a three-layer MLP architecture for binary prediction tasks. The model learns to estimate answerability probability $p(y|H)$ based on fused context-question representations $H \in \mathbb{R}^{6 \times d}$, with positive label ($y=1$) indicating answerable pairs and negative label ($y=0$) otherwise. The optimization objective minimizes the binary cross-entropy loss:
\begin{align}
\mathcal{L} &= -\frac{1}{N} \sum_{i=1}^N \bigl[ y_i \log \sigma(h_\theta(H_i)) \nonumber \\
&\quad + (1-y_i) \log (1-\sigma(h_\theta(H_i))) \bigr],
\end{align}
where $N \in \mathbb{N}^+$ presents total training instances, $y_i \in \{0,1\}$ denotes ground-truth label for $i$-th sample, $h_\theta: \mathbb{R}^{6 \times d} \rightarrow [0,1]^2$ presents MLP with sigmoid activation $\sigma(\cdot)$, and $H_i \in \mathbb{R}^{6 \times d}$ denotes concatenated feature matrix for $i$-th input.

\noindent\textbf{Chunk Selection.} For processed context sequence $[\text{initial}, c_0, c_1, ..., c_m, \text{question}]$, each context chunk $c_i$ is paired with the question component to form context-question pair $X_i = [c_i; \text{question}] \in \mathbb{R}^{(|c_i|+|\text{question}|) \times d}$. Then use the above method to generate the classifier input $H_i \in \mathbb{R}^{6 \times d}$. Through the classifier $h_\theta: \mathbb{R}^{6 \times d} \rightarrow [0,1]^2$, we obtain class-conditional probabilities $\mathbf{p}_i = [T_i, F_i]$ through sigmoid-activated prediction heads, where:
\begin{align}
T_i &= P(y=1|X_i) = \sigma(h_\theta(X_i)_0), \\
F_i &= P(y=0|X_i) = \sigma(h_\theta(X_i)_1).
\end{align}
The relevance score set $\mathbb{T} = \{T_i\}_{i=0}^m$ is aggregated for chunk selection. The compression ratio $\alpha_c \in (0,1]$ is dynamically determined by:
\begin{align}
\alpha_c =  \frac{l_C}{l_T}  \quad (l_C = \sum_{i=0}^m |c_i|, \ l_T \leq L_{\text{max}}),
\end{align}
where $L_{\text{max}}$ denotes the LLM's context window limit and $l_T$ denotes the target context length. The selection criterion retains the top-$\lfloor m/\alpha \rfloor$ chunks $\{c_j\}$ with maximal $T_j$ values. The final compressed context is constructed as:
\begin{align}
H_{\text{comp}} = [\text{initial}; \{c_j\}_{j \in \text{top-}k}; \text{question}] \notag
\\ \quad (k = \lfloor m/\alpha \rfloor),
\end{align}
which preserves original structural components while satisfying $|H_{\text{comp}}| \leq L_{\text{max}}$.

\noindent\textbf{LLM Outputs.}
Subsequently, the compressed input is fed into the backbone LLM. Then the LLM will generate answers to corresponding questions.

\section{Experimental Settings}
\label{sec:exp}

\subsection{Datasets}

We utilize both single-hop and multi-hop QA datasets to collect empirical evidence of our proposed DCS. 

\noindent\textbf{Single-hop QA.} For single-hop QA tasks, the correct answer can be derived by identifying and utilizing a single piece of evidence from the provided context. The datasets include MultiFieldQA\_en \footnote{For this dataset, we adopt two distinct construction methods: one derived from LongBench \cite{bai2024longbench}, and the other from LV-Eval \cite{yuan2024lveval}.} \cite{bai2024longbench,yuan2024lveval}, NarrativeQA \cite{kocisky-etal-2018-narrativeqa}, Qasper \cite{dasigi-etal-2021-dataset}, Loogle-SD \cite{li2023loogle}, and Factrecall \cite{yuan2024lveval}. 
For the datasets MultiFieldQA\_en, Loogle-SD, and Factrecall, we select versions ranging from 16k to 256k tokens.

\noindent\textbf{Multi-hop QA.} For multi-hop QA tasks, accurately deriving an answer requires the integration of multiple pieces of information scattered across different parts of the context. The datasets include HotpotQA \cite{yang-etal-2018-hotpotqa}, 2WikiMQA \cite{ho-etal-2020-constructing}, Musique \cite{trivedi-etal-2022-musique}, Loogle-MR \cite{li2023loogle}, HotpotwikiQA \cite{yuan2024lveval}, and Loogle-CR \cite{li2023loogle}.
For the datasets including Loogle-MR, HotpotwikiQA, and Loogle-CR, we select versions ranging from 16k to 256k tokens.

A more comprehensive introduction to the datasets and tasks is provided in Appendix \ref{benchmarks}.

\subsection{Baselines}
We conduct experiments based on Llama-3-8B-Instruct \cite{grattafiori2024llama3herdmodels}, Mistral-7B-Instruct-V0.1 \cite{jiang2023mistral7b}, and Vicuna-7b-v1.5 \cite{zheng2023judgingllmasajudgemtbenchchatbot} as our backbone LLMs. The maximum length of Llama-3-8B-Instrcut and Mistral-7B-Instruct-v0.1 is 8K and the maximum length of Vicuna-7b-v1.5 is 4K. And we compare our approach with the recent competitive baselines: StreamingLLM \cite{xiao2023efficient}, LM-Infinite \cite{han2024lm}, InfLLM \cite{xiao2024infllm}, and MoICE \footnote{Since it only reported results on Mistral and Llama2, our study follows its setup and compares results only on Mistral and Vicuna (which is based on Llama2).} \cite{lin2024mixtureincontextexpertsenhance}. We adhere to the original settings of all baselines.

\subsection{Hyperparameters}

For Sentence-BERT model, we select paraphrase-multilingual-MiniLM-L12-v2 \cite{DBLP:journals/corr/abs-2002-10957}. More details can be found in Appendix \ref{sentence-transformer}. For percentile-based segmentation threshold $\alpha$, we select 60 for Llama3 and Mistral, and 65 for Vicuna. For the target chunk size, we select 512 for all models. For the target context length, we select 7.5k for Llama3, 7k for Mistral, and 3.5k for Vicuna. The detailed settings of question-aware classifiers can be seen in Table \ref{MLPmodel} in Appendix \ref{MLP Classifier}. The training data is based on AdversarialQA \cite{bartolo2020beat}. More details can be seen in Appendix \ref{AdQA}.

\section{Results}

\begin{table*}[t]
\centering
\footnotesize
\begin{tabular*}{0.99\linewidth}{c|cccccc|c}
\toprule
Single-hop QA & MFQA\_en & Narrativeqa & Qasper & Loogle\_SD & MFQA\_en\_16k & Factrecall\_en & Avg.\\
\midrule
Llama-3-8B-Instruct & 44.30 & 21.54 & \cellcolor[rgb]{.992,.929,.89}\textbf{44.79} & 21.25 & 18.22 & 15.50 & 27.6 \\
with Streaming & 40.04 & 19.30 & 42.52 & 18.51 & 12.84 & 12.36 & 24.26\\
with LM-infinite & 40.08 & 18.83 & 42.53 & 18.20 & 13.45 & 12.16 & 24.20 \\
with Infllm & 44.94 & 19.62 & 44.31 & 19.50 & 15.30 & 19.22 & 27.15\\
with DCS  & \cellcolor[rgb]{.992,.929,.89}\textbf{45.83} & \cellcolor[rgb]{.992,.929,.89}\textbf{23.89} & 44.59 & \cellcolor[rgb]{.992,.929,.89}\textbf{45.10} & \cellcolor[rgb]{.992,.929,.89}\textbf{23.70} & \cellcolor[rgb]{.992,.929,.89}\textbf{29.89} & \cellcolor[rgb]{.992,.929,.89}\textbf{35.50} \\
\midrule
Multi-hop QA & Hotpotqa & 2wikimqa & Musique & Loogle\_MR & Hotpotwikiqa & Loogle\_CR & Avg. \\
\midrule
Llama-3-8B-Instruct & 46.74 & 35.66 & 21.72 & 10.50 & 14.22 & 16.49 & 24.22\\
with Streaming & 43.60 & 35.79 & 18.81 & 9.90 & 12.45 & 14.50 & 22.51\\
with LM-infinite  & 43.85 & 35.79 & 19.87 & 10.96 & 11.98 & 14.26 & 22.79 \\
with Infllm & 47.53 & 35.49 & 24.37 & 10.79 & 7.74 & 15.55 & 23.58\\
with DCS & \cellcolor[rgb]{.992,.929,.89}\textbf{48.81} & \cellcolor[rgb]{.992,.929,.89}\textbf{36.48} & \cellcolor[rgb]{.992,.929,.89}\textbf{28.90} & \cellcolor[rgb]{.992,.929,.89}\textbf{15.10} & \cellcolor[rgb]{.992,.929,.89}\textbf{25.40} & \cellcolor[rgb]{.992,.929,.89}\textbf{19.78} & \cellcolor[rgb]{.992,.929,.89}\textbf{29.07} \\

\bottomrule
\end{tabular*}
\caption{The results on 12 long context reading comprehension datasets based on Llama-3-8B-Instruct. For Loogle\_SD, MFQA\_en\_16k, Factrecall\_en, Loogle\_MR, Hotpotwikiqa, and Loogle\_CR, we select the 16k version for experiments. Best results are bolded. The t-test proves that the improvement is statistically significant ($p < 0.05$). The results based on Mistral and Vicuna are presented in Table \ref{Results on single-hop QA} and Table \ref{Results on multi-hop QA} in Appendix.}
\label{Results on Llama3}
\end{table*}

\subsection{Results on Single-hop QA}

The upper half of Table \ref{Results on Llama3} demonstrates that our DCS achieves an average score of 35.50 on Llama3, representing a 28.62\% improvement over the previous best score. In contrast, existing methods often encounter fragmentation issues when processing lengthy texts, resulting in the loss of semantic coherence and key information. Our dynamic chunking strategy effectively addresses these limitations by preserving semantic integrity and focusing on relevant chunks, thereby enhancing overall understanding. These straightforward yet effective modules significantly enhance the robustness and versatility of our approach, making it a reliable solution for single-hop QA tasks. 

The results based on Mistral and Vicuna are presented in Table \ref{Results on single-hop QA} in Appendix, with our approach achieving improvements of 5.8\% on Mistral and 24.9\% on Vicuna.

\subsection{Results on Multi-hop QA}

The lower half of Table \ref{Results on Llama3} underscores the exceptional performance of DCS in multi-hop QA tasks. Specifically, our approach gets an average score of 29.07. And it achieves a 20.02\% improvement in average scores on Llama3 compared to the previous best scores. Current methods often struggle with multi-hop questions due to their inability to effectively integrate information from multiple sources. Our dynamic chunking strategy, combined with a question-aware classifier, overcomes this limitation by accurately identifying and integrating relevant chunks. Our approach significantly enhances the LLMs' capacity to handle complex reasoning tasks, yielding more precise answers and ensuring reliable and consistent performance across a diverse range of multi-hop QA tasks. 

The results for Mistral and Vicuna are presented in Table \ref{Results on multi-hop QA} in Appendix, with respective improvements of 7.6\% and 7.3\%.

\subsection{Results on Longer Datasets}
To rigorously evaluate our approach's long-context capabilities, we conduct evaluations on extended versions of six benchmark datasets (Loogle\_SD, MultifieldQA\_en, Factrecall\_en, Loogle\_MR, Hotpotwikiqa, and Loogle\_CR), spanning context lengths from 16k to 256k tokens.

\begin{figure}[t]
  \centering
  \subfigure[Results on Single-hop QA (Loogle\_SD, MFQA\_en, and Factrecall\_en). The x-axis represents the length of the input context, ranging from 16k to 256k. The y-axis shows the average score of the model across three datasets.]{
		\label{single}
		\includegraphics[width=0.8\linewidth]{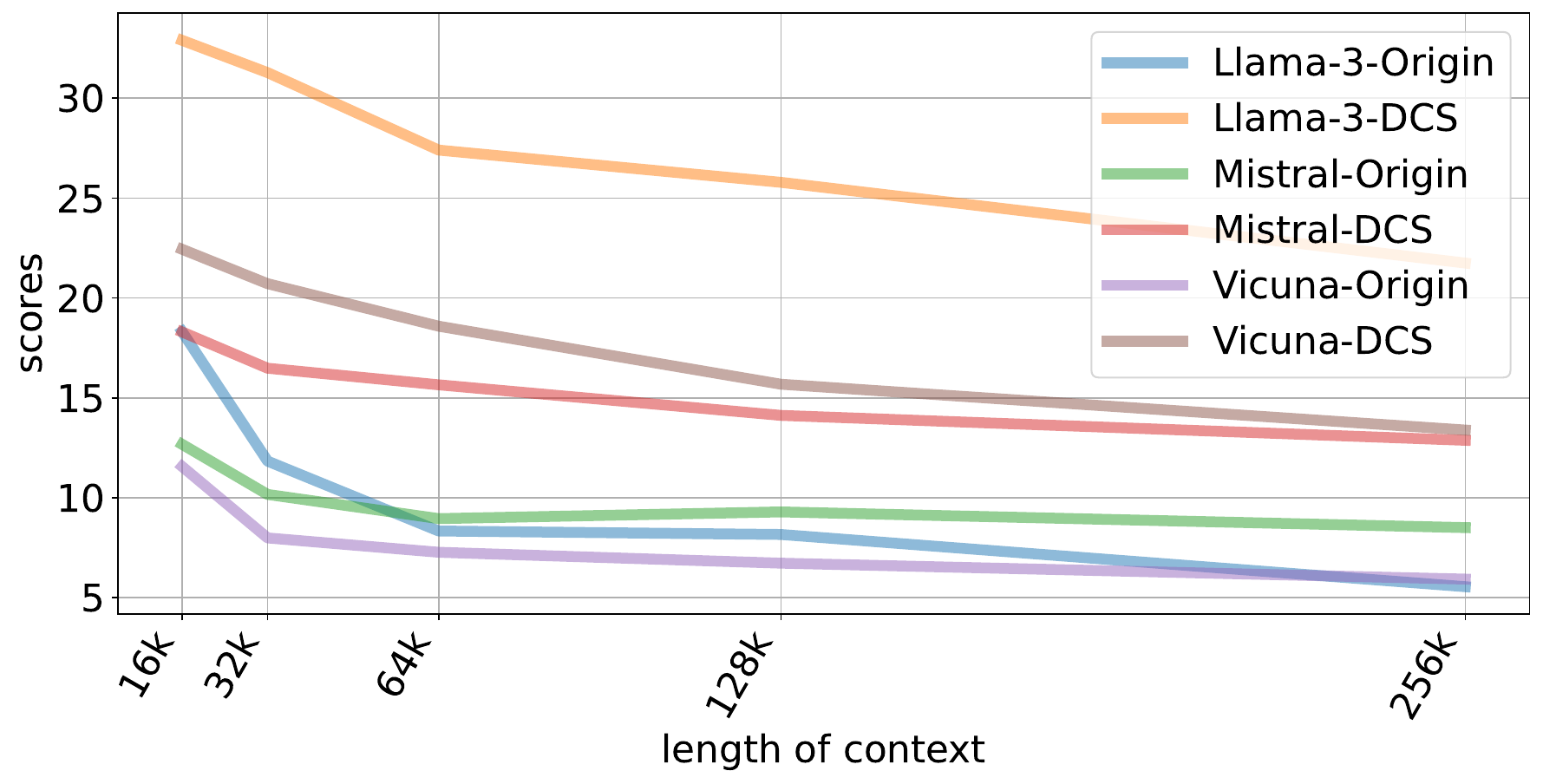}}
  \subfigure[Results on Multi-hop QA (Loogle\_MR, Hotpotwikiqa, and Loogle\_CR). The x-axis represents the length of the input context, ranging from 16k to 256k. The y-axis shows the average score of the model across three datasets.]{
		\label{multi}
		\includegraphics[width=0.8\linewidth]{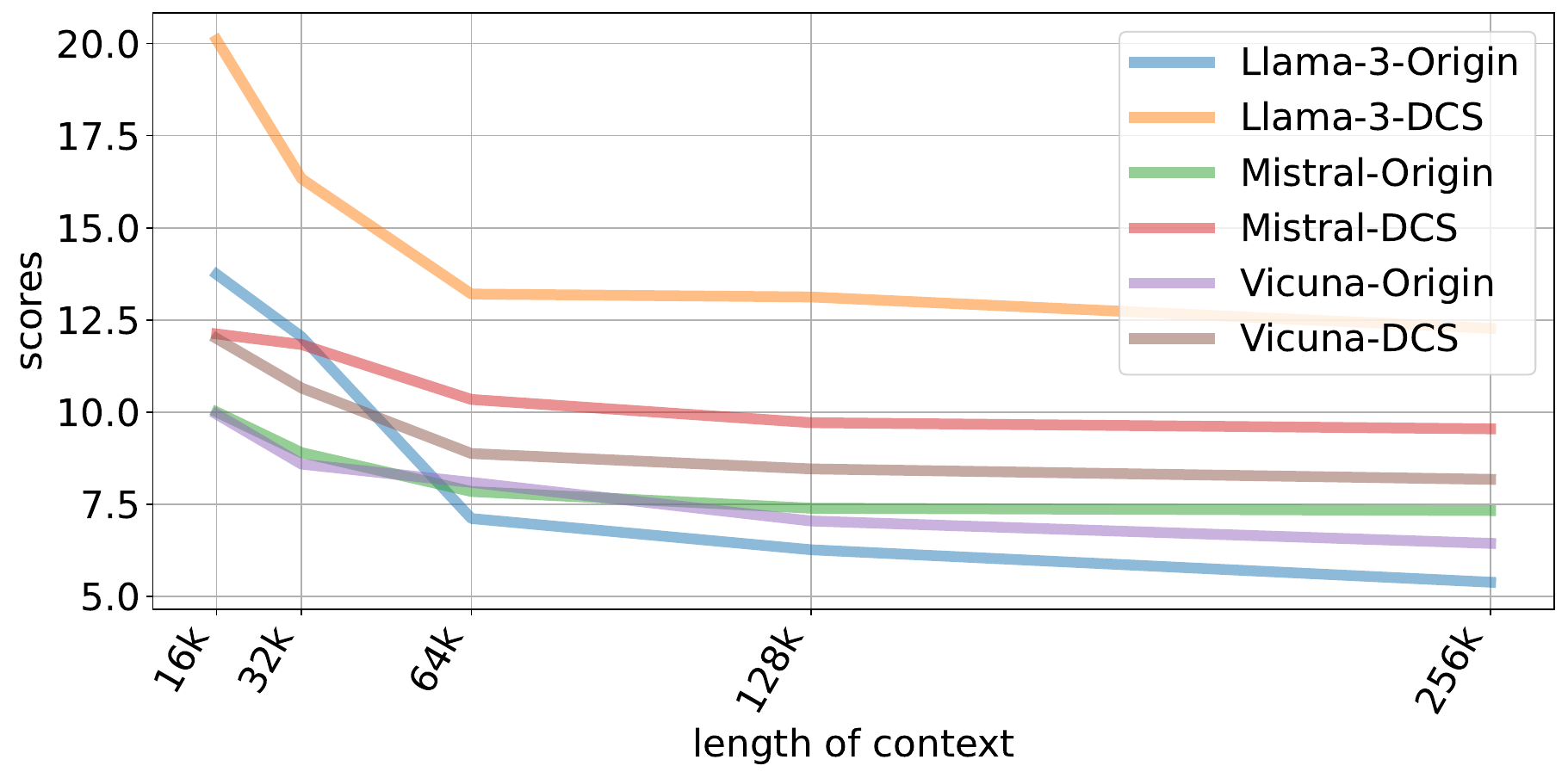}}
  \caption {Results on longer datasets.}
  \label{result:longer}
\end{figure}

As shown in Figure \ref{single} and Figure \ref{multi}, our approach exhibits minimal performance degradation as context lengths increase. In contrast, baselines suffer from significant performance deterioration. This empirical evidence underscores our approach's superior robustness in long-context comprehension tasks. The stability gap widens progressively beyond 64k tokens, where conventional approaches lose critical contextual dependencies. Our approach thus achieves significant improvements in preserving semantic coherence across extended sequences while maintaining robust performance stability.

\subsection{Discussion}

\subsubsection{The Selection of Hyperparameters}

We conduct systematic hyperparameter optimization experiments to identify the optimal configuration. Take Llama3 as an example, specifically, we evaluate different values for chunk length $l$ and percentile-based segmentation threshold $\alpha$. The results in Table \ref{hyperparameters} demonstrate that setting $l=512$ yields the best overall performance. It achieves an average score of 38.08 across all datasets, a significant improvement over alternative configurations (256, 768, or 1024). Similarly, within the tested $\alpha$ range (55-70), $\alpha$=60 produces the highest average score (38.08) while maintaining consistent performance across most datasets. Based on these findings, we establish $l=512$ and $\alpha$=60 as the model's default hyperparameters. This configuration not only delivers optimal performance as measured by the evaluation metrics but also exhibits strong robustness across varying parameter settings.

\begin{table*}[t]
\centering
\footnotesize
\begin{tabular*}{0.90\linewidth}{c|cccccc|c}
\toprule
Chunk Length (*l*) & NarrativeQA & HotpotQA & 2WikiMQA & MFQA\_en & Qasper & Musique & Avg \\
\midrule
256 & 22.22 & 46.42 & \cellcolor[rgb]{.992,.929,.89}\textbf{36.61} & 42.96 & 44.17 & \cellcolor[rgb]{.992,.929,.89}\textbf{31.57} & 37.33 \\
Our (512) & \cellcolor[rgb]{.992,.929,.89}\textbf{23.89} & 48.81 & 36.48 & \cellcolor[rgb]{.992,.929,.89}\textbf{45.83} & \cellcolor[rgb]{.992,.929,.89}\textbf{44.59} & 28.90 & \cellcolor[rgb]{.992,.929,.89}\textbf{38.08} \\
768 & 20.71 & \cellcolor[rgb]{.992,.929,.89}\textbf{50.82} & 35.58 & 45.06  & 43.97  & 25.10 & 36.87 \\
1024 & 20.75 & 48.07 & 35.64 & 44.96  & 44.13  & 26.75 & 36.72 \\
\midrule
Threshold (α) & NarrativeQA & HotpotQA & 2WikiMQA & MFQA\_en & Qasper & Musique & Avg \\
\midrule
55 & 20.41 & 48.33 & \cellcolor[rgb]{.992,.929,.89}\textbf{36.95} & \cellcolor[rgb]{.992,.929,.89}\textbf{46.02} & 43.66 & 29.19 & 37.43 \\
Our (60) & \cellcolor[rgb]{.992,.929,.89}\textbf{23.89} & \cellcolor[rgb]{.992,.929,.89}\textbf{48.81} & 36.48 & 45.83 & 44.59 & 28.90 & \cellcolor[rgb]{.992,.929,.89}\textbf{38.08} \\
65 & 23.06 & 47.34 & 36.35 & 45.07 & \cellcolor[rgb]{.992,.929,.89}\textbf{44.64} & \cellcolor[rgb]{.992,.929,.89}\textbf{29.55} & 37.67 \\
70 & 21.83 & 47.79 & 36.87 & 44.40 & 43.91 & 27.01 & 36.97 \\
\bottomrule
\end{tabular*}
\caption{The results of the selection of hyperparameters on 6 long context reading comprehension datasets based on Llama-3-8B-Instruct. Best results are bolded.}
\label{hyperparameters}
\end{table*}

\subsubsection{Ablation Studies}

We conduct systematic ablation studies to compare dynamic chunking (DC) with fixed chunking (FC) across three base LLMs. As shown in Table \ref{dynamicvsnodynamic}, DC consistently outperforms fixed chunking, achieving average performance gains of 1.12-1.54\% across all LLM-task combinations. These results confirm that our dynamic chunking, through its context-aware optimization, surpasses fixed segmentation approaches. The evidence strongly supports DC's effectiveness in preserving semantic continuity across chunk-level contexts.

\begin{table}[t]
\centering
\small
\begin{tabular*}{\linewidth}{c|ccc}
\toprule
~ &Single-hop QA & Multi-hop QA & Avg. \\
\midrule
Llama3-8B & 36.87 & 34.71 & 35.78 \\
w/ DC & \cellcolor[rgb]{.992,.929,.89}\textbf{38.10} & \cellcolor[rgb]{.992,.929,.89}\textbf{38.06} & \cellcolor[rgb]{.992,.929,.89}\textbf{38.08} \\
w/ FC & 36.66 & 37.26 & 36.96 \\
\midrule
Mistral-7B & 30.63 & 25.01 & 27.82 \\
w/ DC & 30.52 & \cellcolor[rgb]{.992,.929,.89}\textbf{28.79} & \cellcolor[rgb]{.992,.929,.89}\textbf{29.65} \\
w/ FC & \cellcolor[rgb]{.992,.929,.89}\textbf{30.54} & 27.44 & 28.98 \\
\midrule
Vicuna-7B & 25.52 & 15.47 & 20.50 \\
w/ DC & \cellcolor[rgb]{.992,.929,.89}\textbf{26.51} & \cellcolor[rgb]{.992,.929,.89}\textbf{17.34} & \cellcolor[rgb]{.992,.929,.89}\textbf{21.93}\\
w/ FC & 25.43 & 16.39 & 20.91 \\
\bottomrule
\end{tabular*}
\caption{A comparison of average results among the original LLM, dynamic chunking method (w/ DC), and fixed chunking method (w/ FC) on the single-hop QA (Multifieldqa\_en, Narrativeqa and Qasper) and Multi-hop QA (Hotpotqa, 2wikimqa and Musique). Best results are bolded.}
\label{dynamicvsnodynamic}
\end{table}

We also compare our MLP-based question-aware chunk selection method with a cosine similarity (CS) selection approach. As shown in Table \ref{mlpvscos}, the question-aware classifier consistently outperforms the CS across most LLMs and tasks, achieving significant performance improvements. These results highlight the critical role of the question-aware classifier in chunk selection. The ability of the question-aware classifier to capture nonlinear feature interactions is crucial to our approach's ability to make informed chunk selections.

\begin{table}[t]
\centering
\small
\begin{tabular*}{\linewidth}{c|ccc}
\toprule
~ &Single-hop QA & Multi-hop QA & Avg. \\
\midrule
Llama3-8B & 18.32 & 13.74 & 16.03 \\
w/ Classifier & 32.90 & \cellcolor[rgb]{.992,.929,.89}\textbf{20.36} & \cellcolor[rgb]{.992,.929,.89}\textbf{26.85} \\
w/ CS & \cellcolor[rgb]{.992,.929,.89}\textbf{33.07} & 18.60 & 25.84 \\
\midrule
Mistral-7B & 12.68 & 10.00 & 11.34 \\
w/ Classifier & \cellcolor[rgb]{.992,.929,.89}\textbf{18.30} & \cellcolor[rgb]{.992,.929,.89}\textbf{12.38} & \cellcolor[rgb]{.992,.929,.89}\textbf{15.34} \\
w/ CS & 16.16 & 12.00 & 14.08 \\
\midrule
Vicuna-7B & 11.57 & 9.94 & 9.94 \\
w/ Classifier & \cellcolor[rgb]{.992,.929,.89}\textbf{22.44} & \cellcolor[rgb]{.992,.929,.89}\textbf{12.00} & \cellcolor[rgb]{.992,.929,.89}\textbf{17.22}\\
w/ CS & 19.81 & 11.29 & 15.55 \\
\bottomrule
\end{tabular*}
\caption{A comparison of average results among the original model, question-aware classifier method, and cosine similarity method on the single-hop QA (Loogle\_SD, Multifieldqa\_en\_16k and Factrecall\_en) and Multi-hop QA (Loogle\_MIR, Hotpotwikiqa and Loogle\_CR). CS means cosine similarity. Best results are bolded. }
\label{mlpvscos}
\end{table}

\subsubsection{Classifier Robustness to Training Data}
To rigorously assess the stability of our question-aware classifier across diverse training data, we conduct extensive experiments based on three benchmark datasets: AdversarialQA, CoQA \cite{reddy2019coqaconversationalquestionanswering}, and SQuAD \cite{rajpurkar2018knowdontknowunanswerable}. These datasets, which are well-established in the field, provide a robust basis for evaluation. All experiments adhere to the consistent data processing protocols detailed in our methodology section. As shown in Table \ref{mlptrain}, the question-aware classifier exhibits stable performance across different training datasets when evaluated on three backbone LLMs. These results affirm the robust stability of our question-aware classifier's architecture.

\begin{table}[!ht]
\centering
\small
\begin{tabular*}{0.85\linewidth}{c|ccc}
\toprule
~ &SHQA & MHQA & Avg. \\
\midrule
~ & \multicolumn{3}{c}{Llama-3-8B-Instruct} \\
\midrule
w/ AdversarialQA & 38.10 & 38.06 & 38.08 \\
w/ CoQA & 38.01 & 38.11 & 38.06 \\
w/ Squad & 38.09 & 37.78 & 37.93 \\
\midrule
~ & \multicolumn{3}{c}{Mistral-7B-Instruct} \\
\midrule
w/ AdversarialQA & 30.52 & 28.79 & 29.65 \\
w/ CoQA & 30.46 & 27.62 & 29.04 \\
w/ Squad & 30.59 & 28.47 & 29.53 \\
\midrule
~ & \multicolumn{3}{c}{Vicuna-7B} \\
\midrule
w/ AdversarialQA & 26.51 & 17.34 & 21.93\\
w/ CoQA & 26.06 & 16.12 & 21.09 \\
w/ Squad & 26.39 & 17.66 & 22.02 \\
\bottomrule
\end{tabular*}
\caption{A comparison of average results among the question-aware classifier training on different datasets. SHQA represents single-hop QA (Multifieldqa\_en, Narrativeqa and Qasper). MHQA represents multi-hop QA (Hotpotqa, 2wikimqa and Musique).}
\label{mlptrain}
\end{table}

\subsubsection{Latency}
\begin{figure}[t]
  \centering
  \includegraphics[width=0.99\linewidth]{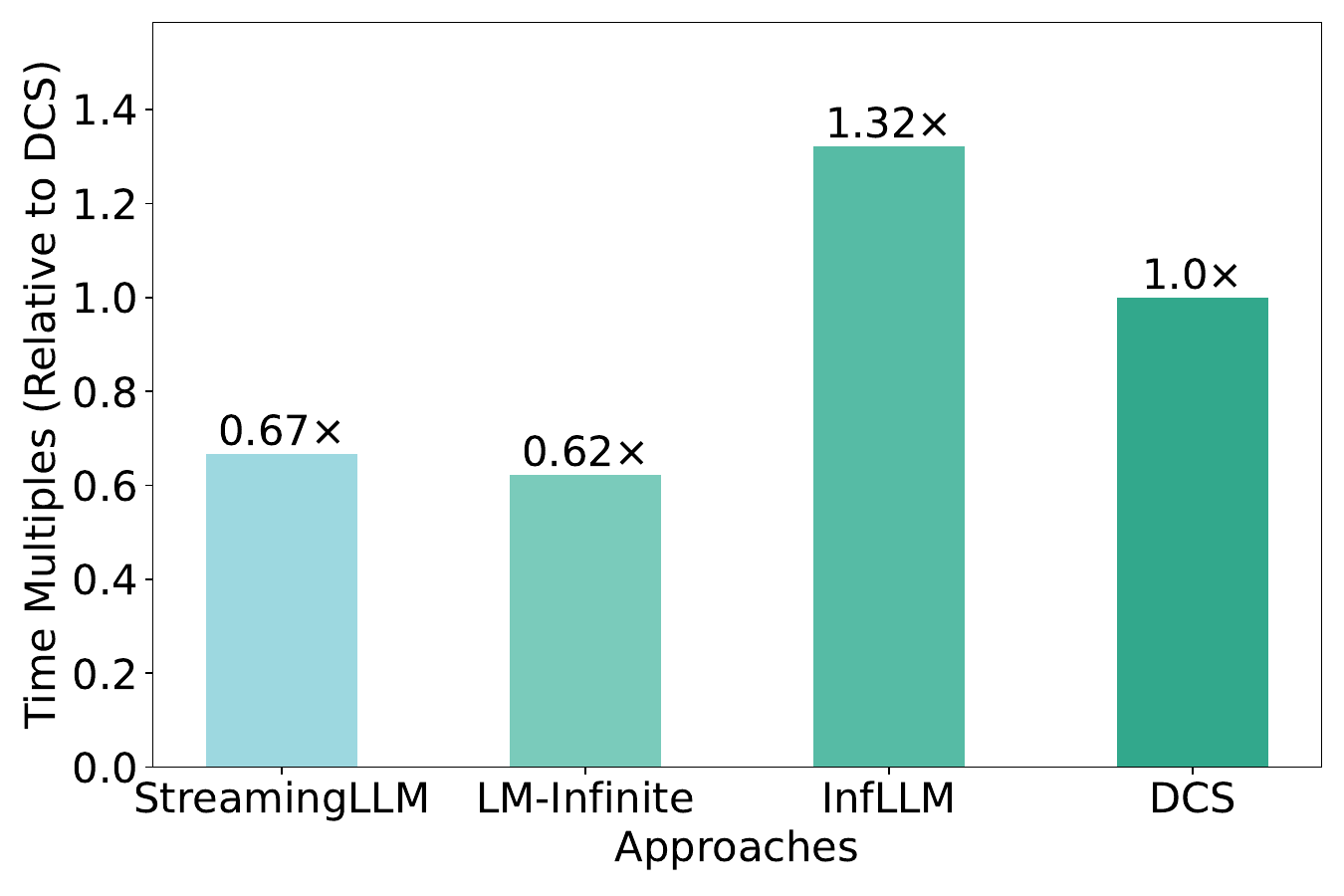}
  \caption {Comparison of time spent by different approaches (relative to our approach).The x-axis represents different approaches. The y-axis is the ratio of time consumption between other methods and our method.}
  \label{latency}
\end{figure}

We conduct comprehensive experiments on the LongBench samples based on Llama-3-8B-Instruct, comparing our approach with Streaming, LM-Infinite, and InfLLM. The latency results can be seen in Figure \ref{latency}. The attention mechanism of LLM typically has a complexity of O($n^2$), which can lead to significant time consumption. In contrast, our approach employs chunking, resulting in a complexity of O($nl$), where n is the sequence length and $l$ is the chunk length (with $l << n$). This complexity is comparable to other long-text approaches, meaning that our approach has limited impact on latency. 

\section{Conclusion}
This paper proposes a simple yet effective approach to enhance the very long-context reading comprehension capabilities of LLMs. Our approach dynamically segments long context into semantically coherent chunks. Then it includes a question-aware classifier to select crucial chunks. Finally, these selected chunks are then concatenated in their original order to fit within the pre-trained context window constraints of the backbone LLMs. Experimental results demonstrate consistent performance improvements across various backbone LLMs when applying our approach. It not only outperforms SOTA methods in terms of average scores but also achieves top rankings across multiple datasets. Notably, it exhibits exceptional robustness, maintaining stable performance despite variations in input length and changes in the training data configuration of the question-aware classifier.

\section{Limitations}
The DCS proposed in this paper primarily addresses long text reading comprehension tasks. However, further exploration of other long text applications warrants more research. Due to limitations in computing resources, this study focuses on only three backbone LLMs and twelve QA datasets. Future experiments could involve additional large models and diverse scenarios to better validate the effectiveness of the proposed DCS. Furthermore, directly applying the modules within the DCS to existing chunk-based methods may yield valuable insights into both the task and the methodology.

\section{Ethics Statement}
The research presented in this paper is founded on open-source LLMs and utilizes publicly available datasets. Consequently, we do not anticipate that our study will have any direct adverse effects. However, it is crucial to recognize that any generative AI technology, including the contributions of our research, must be implemented with caution to avert potentially harmful outcomes.

\newpage
\bibliography{custom}

\newpage
\appendix

\section*{Appendix}

\section{Notation Table}
\label{Notation Table section}
We place a notation table in Table \ref{Notation Table} to facilitate reader reference.

\begin{table}
\centering
\small
\begin{tabularx}{\linewidth}{>{\hsize=0.3333\linewidth\centering\arraybackslash}X|>{\arraybackslash}X}
\toprule
Symbol & Meaning \\
\midrule
$C$ & input context \\
\midrule
$s_i$ & sentences of the input context \\
\midrule
$s_i'$ & merged sentences of the input context \\
\midrule
$l$ & predefined chunk length parameter \\
\midrule
$e_i$ & embedding of merged sentences \\
\midrule
$sim(i, i + 1)$ & the cosine similarity between the $i$-th sentence and the $(i+1)$-th sentence \\
\midrule
$dis(i)$ & the cosine distance between the $i$-th sentence and the $(i+1)$-th sentence \\
\midrule
$\alpha$ & percentile-based segmentation threshold \\
\midrule
$\mathcal{K}$ & positions of maximal dissimilarity indices as segmentation boundaries \\
\midrule
$c^{(i)}_j$ & the $i$-th iteration’s output chunk \\
\midrule
$\mathcal{C}$ & the $i$-th iteration’s output \\
\midrule
$Q$ & input question \\
\midrule
$X$ & context and question tokens sequence \\
\midrule
$C_i$ & tokens of context \\
\midrule
$Q_i$ & tokens of question \\
\midrule
$H$ & hidden states sequence \\
\midrule
$h^{(d)}_i$ & the $i$-th token's hidden state of the $d$-th layer \\
\midrule
$\mathcal{A}$ & the attention scores \\
\midrule
$H_b$ & boundary tokens' hidden state \\
\midrule
$\mathcal{A}_h$ & the attention scores are averaged along the head dimension \\
\midrule
$\mathcal{A}_{QC}$ & cross-attention between question tokens and context tokens\\
\midrule
$\mathcal{A}_{QQ}$ & intra-attention within question tokens \\
\midrule
$\mathcal{A}_{QQ}$ & intra-attention within context tokens \\
\midrule
$\mathbf{a}_C$ & column-wise mean pooling result of cross-attention between question tokens and context tokens \\
\midrule
$\mathbf{a}_Q$ & column-wise mean pooling result of intra-attention within context tokens \\
\midrule
$h^{(d)}_C$ & context-specific representations \\
\midrule
$h^{(d)}_Q$ & question-specific representations \\
\midrule
$\mathcal{L}$ & binary cross-entropy loss of the classifier \\
\midrule
$\alpha_c$ & compression ratio \\
\midrule
$l_C$ & length of context \\
\midrule
$l_T$ & the target context length \\
\midrule
$L_{max}$ & the LLM’s context window limit \\
\midrule
$H_{comp}$ & the final compressed context \\
\bottomrule
\end{tabularx}
\caption{Notation Table}
\label{Notation Table}
\end{table}

\section{Benchmarks}
\label{benchmarks}

\subsection{LongBench}
LongBench is introduced as the pioneering bilingual, multi-task benchmark specifically designed to evaluate long context understanding in LLMs. This benchmark provides a rigorous assessment platform for tasks involving longer sequence inputs that exceed the typical capacity of most language models. LongBench includes 21 datasets, spanning six task categories in both English and Chinese. The average text length is 6,711 words for English and 13,386 characters for Chinese texts. These datasets cover different application areas, including single-document QA, multi-document QA, summarization, few-shot learning, synthetic tasks, and code completion. The inclusion of these diverse and extensive datasets, standardized into a unified format, facilitates automatic evaluation of LLMs' performance in processing and comprehending lengthy textual content. 

In our paper, we choose 6 datasets from single-document QA and multi-document QA. The length of datasets can be seen in Table \ref{lbavg}. The prompts of each dataset can be seen in Figure \ref{lbp} and Figure \ref{lbp2}.

\begin{table}[t]
\centering
\small
\begin{tabular*}{0.85\linewidth}{c|c|c|c}
\toprule
~ & Llama3 & Mistral & Vicuna \\
\midrule
Multifieldqa\_en & 6939 & 7908 & 8116 \\
Narrativeqa & 29869 & 35298 & 36038 \\
Qasper & 5088 & 5693 & 5781 \\
Hotpotqa & 12854 & 14976 & 15331 \\
2wikimqa & 7168 & 8365 & 8485 \\
Musique & 15617 & 18149 & 18556 \\
\bottomrule
\end{tabular*}
\caption{The average number of tokens in the datasets across three different models.}
\label{lbavg}
\end{table}

\begin{figure}[t]
  \centering
  \begin{tcolorbox} 
\textbf{Multifieldqa\_en:} Read the following text and answer briefly. \{context\} Now, answer the following question based on the above text, only give me the answer and do not output any other words. Question: \{input\} Answer:

\textbf{Narrativeqa:} You are given a story, which can be either a novel or a movie script, and a question. Answer the question asconcisely as you can, using a single phrase if possible. Do not provide any explanation. Story: \{context\} Now, answer the question based on the story asconcisely as you can, using a single phrase if possible. Do not provide any explanation. Question: \{input\} Answer:

\textbf{Qasper:} You are given a scientific article and a question. Answer the question as concisely as you can, using a single phrase or sentence if possible. If the question cannot be answered based on the information in the article, write "unanswerable". If the question is a yes/no question, answer "yes", "no", or "unanswerable". Do not provide any explanation. Article: \{context\} Answer the question based on the above article as concisely as you can, using a single phrase or sentence if possible. If the question cannot be answered based on the information in the article, write "unanswerable". If the question is a yes/no question, answer "yes", "no", or "unanswerable". Do not provide any explanation. Question: \{input\} Answer:

    \end{tcolorbox}
  \caption {Prompts of Multifieldqa\_en,Narrativeqa, and Qasper.}
  \label{lbp}
\end{figure}

\begin{figure}[t]
  \centering
  \begin{tcolorbox} 
  
\textbf{Hotpotqa:} Answer the question based on the given passages. Only give me the answer and do not output any other words. The following are given passages.\{context\} Answer the question based on the given passages. Only give me the answer and do not output any other words. Question: \{input\} Answer:

\textbf{2wikimqa:} Answer the question based on the given passages. Only give me the answer and do not output any other words. The following are given passages. \{context\} Answer the question based on the given passages. Only give me the answer and do not output any other words. Question: \{input\} Answer:

\textbf{Musique:} Answer the question based on the given passages. Only give me the answer and do not output any other words. The following are given passages.\{context\} Answer the question based on the given passages. Only give me the answer and do not output any other words. Question: \{input\} Answer:

    \end{tcolorbox}
  \caption {Prompts of Hotpotqa, 2wikimqa, and Musique.}
  \label{lbp2}
\end{figure}

\subsection{LVEval}
LV-Eval is introduced as a sophisticated long-context benchmark designed to address the limitations of existing mainstream benchmarks. This new benchmark challenges state-of-the-art LLMs by featuring five length levels—16k, 32k, 64k, 128k, and 256k words—culminating in an unprecedented context length of 256k words. LV-Eval encompasses two primary tasks: single-hop QA and multi-hop QA, which together include 11 datasets in English or Chinese. To enhance its robustness and fairness, the design of this benchmark incorporates three critical techniques. First, it inserts confusing facts to test models' discernment abilities. Second, it replaces keywords and phrases to challenge model comprehension. Third, it develops keyword-recall-based metrics to provide more accurate performance assessments. By providing controllable evaluations across varying context lengths and incorporating challenging test instances with misleading information, LV-Eval mitigates issues of knowledge leakage and facilitates more objective evaluations of LLMs. Furthermore, LV-Eval highlights concerns about evaluation biases due to knowledge leakage and inaccurate metrics, demonstrating how these issues are effectively reduced within its framework.

In our paper, we choose 6 English datasets. The length of datasets can be seen in Table \ref{lvavg}. The prompts of each dataset can be seen in Figure \ref{lvp} and Figure \ref{lvp2}.

\begin{table}[t]
\centering
\small
\begin{tabular*}{0.75\linewidth}{c|c|c|c}
\toprule
~ & Llama3 & Mistral & Vicuna \\
\midrule
16k & 108100 & 108118 & 108272 \\
32k & 194643 & 194661 & 194815 \\
64k & 365083 & 365101 & 365255 \\
128k & 695415 & 695436 & 695590 \\
256k & 1351528 & 1351546 & 1351700 \\
\bottomrule
\end{tabular*}
\caption{The average number of tokens in different length of datasets across three different models.}
\label{lvavg}
\end{table}

\begin{figure}[t]
  \centering
  \begin{tcolorbox} 
  
\textbf{Loogle\_SD:} Please answer the following question based on the given passages. Questions and answers are only relevant to one passage. Only give me the answer and do not output any other explanation and evidence. Article: \{context\} Please answer the following question based on the above passages. Questions and answers are only relevant to one passage. Only give me the answer and do not output any other explanation and evidence. Question: \{input\} Answer:

\textbf{Multifieldqa\_en:} Please answer the following question based on the given passages. Questions and answers are only relevant to one passage. Only give me the answer and do not output any other explanation and evidence. Article: \{context\} Please answer the following question based on the above passages. Questions and answers are only relevant to one passage. Only give me the answer and do not output any other explanation and evidence. Question: \{input\} Answer:

\textbf{Factrecall\_en:} Please answer the following questions based on the given article. Article: \{context\} Please answer the following questions based on the above article. Question: \{input\} Answer:

    \end{tcolorbox}
  \caption {Prompts of Loogle\_SD, Multifieldqa\_en, and Factrecall\_en.}
  \label{lvp}
\end{figure}

\begin{figure}[t]
  \centering
  \begin{tcolorbox} 

\textbf{Loogle\_MR:} Please answer the following question based on the given passages. Questions and answers are only relevant to one passage. Only give me the answer and do not output any other explanation and evidence. Article: \{context\} Please answer the following question based on the above passages. Questions and answers are only relevant to one passage. Only give me the answer and do not output any other explanation and evidence. Question: \{input\} Answer:

\textbf{Hotpotwikiqa:} Answer the question based on the given passages. Questions and answers are only relevant to some passages. Only give me the answer and do not output any other explanation and evidence. Article: \{context\} Please answer the following question based on the above passages. Questions and answers are only relevant to some passages. Only give me the answer and do not output any other explanation and evidence. Question: \{input\} Answer:

\textbf{Loogle\_CR:} Please answer the following question based on the given passages. Questions and answers are only relevant to one passage. Only give me the answer and do not output any other explanation and evidence. Article: \{context\} Please answer the following question based on the above passages. Questions and answers are only relevant to one passage. Only give me the answer and do not output any other explanation and evidence. Question: \{input\} Answer:

    \end{tcolorbox}
  \caption {Prompts of Loogle\_SD, Multifieldqa\_en, and Factrecall\_en.}
  \label{lvp2}
\end{figure}

\subsection{More Results}
\subsubsection{Results on Single-hop QA}
The lower portion of Table \ref{Results on single-hop QA} highlights the significant improvements achieved by our DCS approach on the Mistral and Vicuna models. For the Mistral-7B-Instruct model, DCS attains an average score of 24.42, outperforming other methods. MoICE achieves strong results with scores of 44.39 on MFQA\_en and 30.89 on Qasper. However, DCS surpasses it on average, demonstrating its stability and versatility.
Similarly, for the Vicuna-7B model, DCS exhibits superior performance with an average score of 24.48. MoICE performs well on MFQA\_en (42.29) and Loogle\_SD (14.63), while Infllm shows strength in Qasper (24.35) and Factrecall\_en (16.65). Despite these strong performances, DCS provides a more balanced and enhanced performance across all datasets.
These results underscore the efficacy of the DCS approach in bolstering the robustness and adaptability of LLMs for single-hop QA tasks.

\begin{table*}[t]
\centering
\footnotesize
\begin{tabular*}{0.98\linewidth}{c|cccccc|c}
\toprule
Model & MFQA\_en & Narrativeqa & Qasper & Loogle\_SD & MFQA\_en\_16k & Factrecall\_en & Avg.\\
\midrule
Llama-3-8B-Instruct & 44.30 & 21.54 & \cellcolor[rgb]{.992,.929,.89}\textbf{44.79} & 21.25 & 18.22 & 15.50 & 27.6 \\
with Streaming & 40.04 & 19.30 & 42.52 & 18.51 & 12.84 & 12.36 & 24.26\\
with LM-infinite & 40.08 & 18.83 & 42.53 & 18.20 & 13.45 & 12.16 & 24.20 \\
with Infllm & 44.94 & 19.62 & 44.31 & 19.50 & 15.30 & 19.22 & 27.15\\
with DCS  & \cellcolor[rgb]{.992,.929,.89}\textbf{45.83} & \cellcolor[rgb]{.992,.929,.89}\textbf{23.89} & 44.59 & \cellcolor[rgb]{.992,.929,.89}\textbf{45.10} & \cellcolor[rgb]{.992,.929,.89}\textbf{23.70} & \cellcolor[rgb]{.992,.929,.89}\textbf{29.89} & \cellcolor[rgb]{.992,.929,.89}\textbf{35.50} \\
\midrule
Mistral-7B-Instruct  & 40.81 & \cellcolor[rgb]{.992,.929,.89}\textbf{20.89} & 30.19 & 19.13 & 16.62 & 2.29 & 21.66\\
with Streaming & 33.87 & 12.60 & 17.19 & 11.80 & 14.18 & 29.64 & 19.88\\
with LM-infinite & 34.23 & 12.87 & 17.30 & 12.06 & 14.10 & \cellcolor[rgb]{.992,.929,.89}\textbf{31.36} & 20.32 \\
with Infllm & 42.66 & 14.59 & 22.08 & 18.15 & 16.27 & 24.64 & 23.07 \\
with MoICE & \cellcolor[rgb]{.992,.929,.89}\textbf{44.39} & 17.03 & \cellcolor[rgb]{.992,.929,.89}\textbf{30.89} & 20.81 & 16.62 & 2.64 & 22.06 \\
with DCS & 42.31 & 18.63 & 30.64 & \cellcolor[rgb]{.992,.929,.89}\textbf{24.51} & \cellcolor[rgb]{.992,.929,.89}\textbf{23.76} & 6.64 & \cellcolor[rgb]{.992,.929,.89}\textbf{24.42}\\
\midrule
Vicuna-7B & 38.24 & 14.95 & 23.38 & 14.11 & 13.79 & 6.81 & 18.55\\
with Streaming & 32.67 & 15.37 & 23.38 & 13.11 & 13.82 & 2.74 & 16.85\\
with LM-Infinite & 32.30 & 14.12 & 22.94 & 13.68 & 13.84 & 3.30 & 16.70\\
with InfLLM & 37.16 & \cellcolor[rgb]{.992,.929,.89}\textbf{16.07} & \cellcolor[rgb]{.992,.929,.89}\textbf{24.35} & 11.29 & 5.92 & 16.65 & 18.57\\
with MoICE & \cellcolor[rgb]{.992,.929,.89}\textbf{42.29} & 14.84 & 23.30 & 14.63 & 14.23 & 8.27 & 19.59 \\
with DCS & 40.13 & 15.60 & 23.81 & \cellcolor[rgb]{.992,.929,.89}\textbf{20.19} & \cellcolor[rgb]{.992,.929,.89}\textbf{19.87} & \cellcolor[rgb]{.992,.929,.89}\textbf{27.26} & \cellcolor[rgb]{.992,.929,.89}\textbf{24.48} \\
\bottomrule
\end{tabular*}
\caption{Results on single-hop QA}
\label{Results on single-hop QA}
\end{table*}

\subsubsection{Results on Multi-hop QA}
The lower portion of Table \ref{Results on multi-hop QA} highlights the outstanding performance of our DCS approach in multi-hop QA tasks for the Mistral and Vicuna models. For the Mistral-7B-Instruct model, DCS achieves an average score of 20.59, representing a substantial improvement over other methods. MoICE performs well, scoring 30.18 on Hotpotqa and 20.87 on Loogle\_CR. However, DCS consistently outperforms it across multiple datasets, significantly enhancing the model's ability to handle complex reasoning tasks.
Similarly, for the Vicuna-7B model, DCS demonstrates superior performance with an average score of 14.67, surpassing other methods. InfLLM and MoICE achieve notable results in specific datasets: InfLLM scores 12.64 on Loogle\_MR, and MoICE scores 15.74 on Hotpotwikiqa. Despite these strong performances, DCS maintains a more consistent and enhanced performance across all datasets.
These results underscore the effectiveness of our dynamic chunking strategy combined with a question-aware classifier. This approach overcomes the limitations of current methods that struggle with multi-hop questions.

\begin{table*}[t]
\centering
\footnotesize
\begin{tabular*}{0.95\linewidth}{c|cccccc|c}
\toprule
Model & Hotpotqa & 2wikimqa & Musique & Loogle\_MR & Hotpotwikiqa & Loogle\_CR & Avg. \\
\midrule
Llama-3-8B-Instruct & 46.74 & 35.66 & 21.72 & 10.50 & 14.22 & 16.49 & 24.22\\
with Streaming & 43.60 & 35.79 & 18.81 & 9.90 & 12.45 & 14.50 & 22.51\\
with LM-infinite  & 43.85 & 35.79 & 19.87 & 10.96 & 11.98 & 14.26 & 22.79 \\
with Infllm & 47.53 & 35.49 & 24.37 & 10.79 & 7.74 & 15.55 & 23.58\\
with DCS & \cellcolor[rgb]{.992,.929,.89}\textbf{48.81} & \cellcolor[rgb]{.992,.929,.89}\textbf{36.48} & \cellcolor[rgb]{.992,.929,.89}\textbf{28.90} & \cellcolor[rgb]{.992,.929,.89}\textbf{15.10} & \cellcolor[rgb]{.992,.929,.89}\textbf{25.40} & \cellcolor[rgb]{.992,.929,.89}\textbf{19.78} & \cellcolor[rgb]{.992,.929,.89}\textbf{29.07} \\
\midrule
Mistral-7B-Instruct & 36.89 & 26.71 & 11.42 & 9.47 & 6.07 & 14.47 & 17.51\\
with Streaming  & 23.80 & 19.37 & 5.64 & 7.14 & 5.90 & 10.99 & 12.14 \\
with LM-infinite & 24.85 & 21.63 & 5.12 & 8.47 & 5.78 & 11.39 & 12.87 \\
with Infllm & 28.89 & 24.19 & 12.22 & 9.14 & 7.16 & 13.12 & 15.79\\
with MoICE & 30.18 & 25.72 & 12.95 & \cellcolor[rgb]{.992,.929,.89}\textbf{15.35} & 9.73 & \cellcolor[rgb]{.992,.929,.89}\textbf{20.87} & 19.13 \\
with DCS & \cellcolor[rgb]{.992,.929,.89}\textbf{39.36} & \cellcolor[rgb]{.992,.929,.89}\textbf{28.27} & \cellcolor[rgb]{.992,.929,.89}\textbf{18.75} & 10.59 & \cellcolor[rgb]{.992,.929,.89}\textbf{11.53} & 15.02 & \cellcolor[rgb]{.992,.929,.89}\textbf{20.59}\\
\midrule
Vicuna-7B & 22.02 & 18.02 & 6.38 & 10.61 & 4.32 & 14.90 & 12.71\\
with Streaming & 22.94 & 18.15 & 6.77 & 10.03 & 5.44 & 13.89 & 12.87 \\
with LM-Infinite & 21.80 & 18.12 & 7.29 & 10.17 & 5.46 & 14.57 & 12.91\\
with InfLLM & 23.05 & 17.70 & 4.69 & \cellcolor[rgb]{.992,.929,.89}\textbf{12.64} & 13.81 & 3.99 & 12.65\\
with MoICE & 22.81 & 18.62 & 5.63 & 7.07 & \cellcolor[rgb]{.992,.929,.89}\textbf{15.74} & 12.17 & 13.67 \\
with DCS & \cellcolor[rgb]{.992,.929,.89}\textbf{24.57} & \cellcolor[rgb]{.992,.929,.89}\textbf{19.42} & \cellcolor[rgb]{.992,.929,.89}\textbf{8.04} & 12.52 & 8.33 & \cellcolor[rgb]{.992,.929,.89}\textbf{15.14} & \cellcolor[rgb]{.992,.929,.89}\textbf{14.67} \\
\bottomrule
\end{tabular*}
\caption{Results on multi-hop QA}
\label{Results on multi-hop QA}
\end{table*}

\subsubsection{Results on More Tasks}
Our primary goal is to enhance the long-text reading comprehension capabilities of LLMs, so we utilize widely recognized benchmarks in this area. To validate the generalization ability of our approach, we conduct additional experiments on various tasks. The results in Table \ref{Results on more tasks} show that our approach outperforms other baselines and the original LLM (ori). While domain-specific long-text datasets are currently limited, we recognize their importance and will explore this direction in future work. Our extensive validation in general domains suggests that our approach has the potential to yield strong results in specific domains as well.

\begin{table*}[t]
\centering
\footnotesize
\begin{tabular*}{0.95\linewidth}{c|cccccc}
\toprule
Model & TREC& Multi-News& Passage Retrieval En & Repobench-P& Passkey & Math.Find \\
\midrule
Llama-3-8B-Instruct & 0.00 & \cellcolor[rgb]{.992,.929,.89}\textbf{27.78} & 67.00 & 14.17 & 6.78 & 32.00\\
with Streaming & \cellcolor[rgb]{.992,.929,.89}\textbf{0.50} & 27.73 & 48.50 & 12.16 & 6.78 & 13.43\\
with LM-infinite  & \cellcolor[rgb]{.992,.929,.89}\textbf{0.50} & 27.66 & 48.00 & 11.73 & 6.78 & 12.86 \\
with Infllm & 0.00 & 27.61 & 74.50 & 11.63 & \cellcolor[rgb]{.992,.929,.89}\textbf{100.00} & 16.57\\
with DCS & \cellcolor[rgb]{.992,.929,.89}\textbf{0.50} & 27.73 & \cellcolor[rgb]{.992,.929,.89}\textbf{90.00} & \cellcolor[rgb]{.992,.929,.89}\textbf{15.04} & 82.2 & \cellcolor[rgb]{.992,.929,.89}\textbf{33.14} \\
\bottomrule
\end{tabular*}
\caption{Results on more tasks. TREC is a few-shot question answering task. Multi-News is a multi-document summarization task. And RepoBench-P is a code generation task.}
\label{Results on more tasks}
\end{table*}

\subsubsection{Results on Larger LLM}
To verify the generalizability of our approach, we perform additional validation using the Vicuna-13B model (larger than our original setting). The experimental results in Table \ref{Results on Vicuna-13B} demonstrate that our approach maintains its performance advantage at this scale. This shows our potential on larger LLMs.

\begin{table*}[t]
\centering
\footnotesize
\begin{tabular*}{0.85\linewidth}{c|cccccc|c}
\toprule
Model & Narrativeqa & Hotpotqa & 2wikimqa & MFQA\_en & Qasper & Musique & Avg.\\
\midrule
Vicuna-13B & 15.32 & 33.26 & 29.28 & 42.30 & 23.96 & 14.56 & 26.45 \\
with Streaming & 13.04 & 18.88 & 23.40 & 32.62 & 21.63 & 5.24 & 19.14 \\
with LM-Infinite & 13.67 & 21.68 & 24.31 & 33.04 & 21.58 & 5.47 & 19.94 \\
with InfLLM & \cellcolor[rgb]{.992,.929,.89}\textbf{15.69} & 37.62 & \cellcolor[rgb]{.992,.929,.89}\textbf{33.79} & 39.98 & 26.67 & 12.58 & 27.70 \\
with DCS & 14.60 & \cellcolor[rgb]{.992,.929,.89}\textbf{38.14} & 30.65 & \cellcolor[rgb]{.992,.929,.89}\textbf{46.75} & \cellcolor[rgb]{.992,.929,.89}\textbf{28.35} & \cellcolor[rgb]{.992,.929,.89}\textbf{18.76} & \cellcolor[rgb]{.992,.929,.89}\textbf{29.54} \\

\bottomrule
\end{tabular*}
\caption{Results on Vicuna-13B}
\label{Results on Vicuna-13B}
\end{table*}

\subsubsection{Compare with Overlapping Chunking Method}
We perform controlled experiments comparing our dynamic approach with an overlapping chunking baseline (chunk length = 384 + 128 overlap) on Llama3-8B-Instruct. The results in Table \ref{Results of overlapping} confirm that our approach outperforms this alternative approach.

\begin{table*}[t]
\centering
\footnotesize
\begin{tabular*}{0.85\linewidth}{c|cccccc|c}
\toprule
Model & Narrativeqa & Hotpotqa & 2wikimqa & MFQA\_en & Qasper & Musique & Avg.\\
\midrule
with overlapping & 22.70 & 44.84 & \cellcolor[rgb]{.992,.929,.89}\textbf{40.01} & 43.66 & 43.62 & 26.94 & 36.96 \\ 
with DCS & \cellcolor[rgb]{.992,.929,.89}\textbf{23.89} & \cellcolor[rgb]{.992,.929,.89}\textbf{48.81} & 36.48 & \cellcolor[rgb]{.992,.929,.89}\textbf{45.83} & \cellcolor[rgb]{.992,.929,.89}\textbf{44.59} & \cellcolor[rgb]{.992,.929,.89}\textbf{28.90} & \cellcolor[rgb]{.992,.929,.89}\textbf{38.08} \\

\bottomrule
\end{tabular*}
\caption{Results of comparison with overlapping chunking method on Llama-3}
\label{Results of overlapping}
\end{table*}

\section{Sentence-BERT}
\label{sentence-transformer}
Sentence-BERT is a significant advancement over BERT and RoBERTa, designed to generate semantically meaningful sentence embeddings more efficiently. By leveraging siamese and triplet network structures during fine-tuning, Sentence-BERT enables the encoding of sentences into embeddings that can be compared using simple cosine similarity. This approach dramatically reduces the computational overhead for tasks such as identifying the most similar pair in a collection of sentences—from approximately 65 hours with BERT to about 5 seconds with Sentence-BERT, while maintaining BERT's high accuracy. Evaluated on standard semantic textual similarity (STS) tasks and transfer learning tasks, both Sentence-BERT and its RoBERTa-based variant (SRoBERTa) consistently outperform other state-of-the-art sentence embedding methods.

For our work, we select paraphrase-multilingual-MiniLM-L12-v2. MiniLM is a compact language model derived from larger pre-trained Transformer models, such as BERT, through a process of knowledge distillation. It focuses on deeply mimicking the self-attention modules of the teacher model, particularly those in the final Transformer layer, to ensure efficiency while preserving performance. Unlike previous approaches that perform layer-to-layer distillation, MiniLM's method alleviates the challenge of layer mapping between teacher and student models and offers flexibility in the student model's layer number. Additionally, MiniLM introduces distilling the scaled dot-product between values in the self-attention module as a form of deep self-attention knowledge, alongside traditional attention distributions. This approach allows for relation matrices with consistent dimensions without additional parameters, accommodating arbitrary hidden dimensions in the student model. The use of a teacher assistant further enhances the effectiveness of this distillation process.

\section{Question-aware Classifier}
\label{MLP Classifier}
We selected three datasets as the training sets for the classifier to use in experiments and comparisons, with their specific details shown in Table \ref{details MLPtd}.
The detailed setups of question-aware classifiers can be seen in Table \ref{MLPmodel}.

\begin{table}[t]
\centering
\small
\begin{tabular*}{0.75\linewidth}{c|c|c|c}
\toprule
~ & Train & Valid & Test \\
\midrule
AdversarialQA & 60000 & 6000 & 6000 \\
CoQA & 87418 & 4422 & 4422 \\
Squad & 74896 & 2398 & 2398 \\
\bottomrule
\end{tabular*}
\caption{Details of classifier training data}
\label{details MLPtd}
\end{table}

\begin{figure}[t]
  \centering
  \begin{tcolorbox} 
<|begin\_of\_text|>Beyoncé Giselle Knowles-Carter (/biːˈjɒnseɪ/ bee-YON-say) (born September 4, 1981) is an American singer, songwriter, record producer and actress. ... earned five Grammy Awards and featured the Billboard Hot 100 number-one singles "Crazy in Love" and "Baby Boy".<|begin\_of\_text|>

Now, answer the question based on the story asconcisely as you can, using a single phrase if possible. Do not provide any explanation.

Question: When did Beyonce start becoming popular?

Answer:

    \end{tcolorbox}
  \caption {An example of question-aware classifier input data}
  \label{mctd}
\end{figure}

\begin{table}[ht]
\centering
\small
\begin{tabular*}{0.95\linewidth}{c|c|c|c}
\toprule
~ & Llama3 & Mistral & Vicuna \\
\midrule
\multicolumn{4}{c}{trained on AdversarialQA} \\
\midrule
$W_0$ & 24576*8192 & 24576*4096 & 24576*4096 \\
$W_1$ & 8192*1024 & 4096*256 & 4096*1024 \\
$W_2$ & 1024*2 & 256*2 & 1024*2 \\
Epochs & 20 & 10 & 20 \\
Lr & 1e-5 & 1e-5 & 1.5e-5 \\
\midrule
\multicolumn{4}{c}{trained on CoQA} \\
\midrule
$W_0$ & 24576*4096 & 24576*4096 & 24576*4096 \\
$W_1$ & 4096*256 & 4096*2048 & 4096*4 \\
$W_2$ & 256*2 & 2048*2 & 4*2 \\
Epochs & 20 & 20 & 20 \\
Lr & 2e-5 & 2e-5 & 3e-5 \\
\midrule
\multicolumn{4}{c}{trained on Squad} \\
\midrule
$W_0$ & 24576*4096 & 24576*8192 & 24576*8192 \\
$W_1$ & 4096*512 & 8192*1024 & 8192*128 \\
$W_2$ & 512*2 & 1024*2 & 128*2 \\
Epochs & 10 & 20 & 10 \\
Lr & 1.5e-5 & 1.5e-5 & 1.5e-5 \\
\bottomrule
\end{tabular*}
\caption{Hyperparameters of question-aware classifiers}
\label{MLPmodel}
\end{table}

\subsection{AdversarialQA}
\label{AdQA}
AdversarialQA is a dataset specifically designed to challenge and enhance reading comprehension models by integrating them into the annotation process. In this approach, human annotators craft questions in an adversarial manner, targeting the weaknesses of the reading comprehension (RC) model to generate questions that are particularly difficult to answer correctly. An example of AdversarialQA is illustrated in Figure \ref{adqa}.

\begin{figure}[t]
  \centering
  \begin{tcolorbox} 
\textbf{Context:}
Another approach to brain function is to examine the consequences of damage to specific brain areas. ... In animal studies, most commonly involving rats, it is possible to use electrodes or locally injected chemicals to produce precise patterns of damage and then examine the consequences for behavior.
    \tcblower
\textbf{Question:}
What has been injected into rats to produce precise patterns of damage?

\textbf{Ispossitive:}
True
    \end{tcolorbox}
  \caption {An example of context-question pairs of AdversarialQA}
  \label{adqa}
\end{figure}

\subsection{CoQA}
The CoQA dataset was introduced to drive the development of Conversational question-answering systems, facilitating machines' ability to gather information through natural dialogue. It comprises 127,000 questions and answers derived from 8,000 conversations across seven diverse domains, bridging the gap between human conversation and machine comprehension. The questions in CoQA are designed to reflect conversational patterns, with answers provided in the free-form text and corresponding evidence highlighted in the original passages. A detailed analysis of CoQA reveals that it encompasses complex phenomena such as coreference and pragmatic reasoning, presenting challenges not typically found in traditional reading comprehension datasets. An example of CoQA is illustrated in Figure \ref{coqa}.

\begin{figure}[t]
  \centering
  \begin{tcolorbox} 
\textbf{Context:}
The Vatican Apostolic Library (), more commonly called the Vatican Library or simply the Vat, is the library of the Holy See, located in Vatican City. ... Only a handful of volumes survive from this period, though some are very significant.
    \tcblower
\textbf{Question:}
When was the Vat formally opened?

\textbf{Ispossitive:}
True
    \end{tcolorbox}
  \caption {An example of context-question pairs of CoQA}
  \label{coqa}
\end{figure}

\subsection{Squad}
SQuAD 2.0 is the latest iteration of the Stanford Question Answering Dataset. It addresses limitations in previous extractive reading comprehension systems by incorporating both answerable and unanswerable questions. While earlier datasets focused exclusively on questions with answers present in the context or utilized easily identifiable, automatically generated unanswerable questions, SQuAD 2.0 integrates over 50,000 unanswerable questions crafted adversarially by crowdworkers to closely resemble answerable ones. This new version challenges systems not only to locate correct answers within a context document but also to recognize when a question cannot be answered based on the provided information, thereby requiring them to abstain from guessing. The integration of existing SQuAD data with these carefully designed unanswerable questions makes SQuAD 2.0 a significantly more challenging task for natural language understanding models. An example of SQuAD can be seen in Figure \ref{squad}.

\begin{figure}[t]
  \centering
  \begin{tcolorbox} 
\textbf{Context:}
Beyoncé Giselle Knowles-Carter (/biːˈjɒnseɪ/ bee-YON-say) (born September 4, 1981) is an American singer, songwriter, record producer and actress. ... earned five Grammy Awards and featured the Billboard Hot 100 number-one singles "Crazy in Love" and "Baby Boy".

    \tcblower
\textbf{Question:}
When did Beyonce start becoming popular?

\textbf{Ispossitive:}
True
    \end{tcolorbox}
  \caption {An example of context-question pairs of Squad}
  \label{squad}
\end{figure}

\subsection{Reclor}
ReClor (Reading Comprehension with Logical Reasoning) \cite{yu2020reclorreadingcomprehensiondataset} is a specialized dataset to evaluate machine reading comprehension through logical reasoning tasks. It comprises 6,138 questions extracted from standardized exams such as the GMAT (Graduate Management Admission Test) and LSAT (Law School Admission Test). Each instance includes a context passage, a question, and four answer choices (only one correct). The dataset is partitioned into training (4,638 examples), validation (500 examples), and testing set (1,000 examples). Unlike conventional reading comprehension datasets that may contain redundant information, ReClor uniquely requires rigorous logical reasoning, where every sentence in the passage is semantically critical for deriving the correct answer.

To verify the robustness of the classifier, we conduct further experiments using the ReClor dataset. The results in Table \ref{mlptrain_appendix}, consistently align with our original findings across all three datasets, again highlighting the robustness of our approach.

\begin{table}[t]
\centering
\small
\begin{tabular*}{0.85\linewidth}{c|ccc}
\toprule
~ &SHQA & MHQA & Avg. \\
\midrule
~ & \multicolumn{3}{c}{Llama-3-8B-Instruct} \\
\midrule
w/ AdversarialQA & 38.10 & 38.06 & 38.08 \\
w/ CoQA & 38.01 & 38.11 & 38.06 \\
w/ Squad & 38.09 & 37.78 & 37.93 \\
w/ ReClor & 38.15 & 37.52 & 37.83 \\
\bottomrule
\end{tabular*}
\caption{A comparison of average results among the question-aware classifier training on different datasets. SHQA represents single-hop QA (Multifieldqa\_en, Narrativeqa and Qasper). MHQA represents multi-hop QA (Hotpotqa, 2wikimqa and Musique).}
\label{mlptrain_appendix}
\end{table}

\end{CJK}
\end{document}